\pdfoutput=1

\documentclass[11pt]{article}

\usepackage[final]{acl}

\usepackage{times}
\usepackage{latexsym}

\usepackage[T1]{fontenc}

\usepackage[utf8]{inputenc}

\usepackage{microtype}

\usepackage{inconsolata}

\usepackage{booktabs} 
\usepackage{multirow} 
\usepackage{graphicx} 
\usepackage{amsmath} 
\usepackage{url} 
\usepackage{makecell} 
\usepackage{xcolor} 
 
\definecolor{message_gray}{gray}{0.6} 
\definecolor{message_orange}{HTML}{8E4400} 
\definecolor{message_blue}{HTML}{004A8E} 

\title{Referring Expression Generation in Visually Grounded Dialogue with Discourse-aware Comprehension Guiding}

\author{Bram Willemsen \and Gabriel Skantze \\
        Division of Speech, Music and Hearing \\
        KTH Royal Institute of Technology \\
        Stockholm, Sweden \\
        \texttt{\{bramw,skantze\}@kth.se}}

\begin{document}
\maketitle
\begin{abstract}
We propose an approach to referring expression generation (REG) in visually grounded dialogue that is meant to produce referring expressions (REs) that are both discriminative and discourse-appropriate. 
Our method constitutes a two-stage process. 
First, we model REG as a text- and image-conditioned next-token prediction task. 
REs are autoregressively generated based on their preceding linguistic context and a visual representation of the referent.
Second, we propose the use of discourse-aware comprehension guiding as part of a generate-and-rerank strategy through which candidate REs generated with our REG model are reranked based on their discourse-dependent discriminatory power.
Results from our human evaluation indicate that our proposed two-stage approach is effective in producing discriminative REs, with higher performance in terms of text-image retrieval accuracy for reranked REs compared to those generated using greedy decoding.
\end{abstract}

\section{Introduction}
\label{sec:introduction}
\begin{figure}[t!]
    \centering
    \includegraphics[width=.75\linewidth]{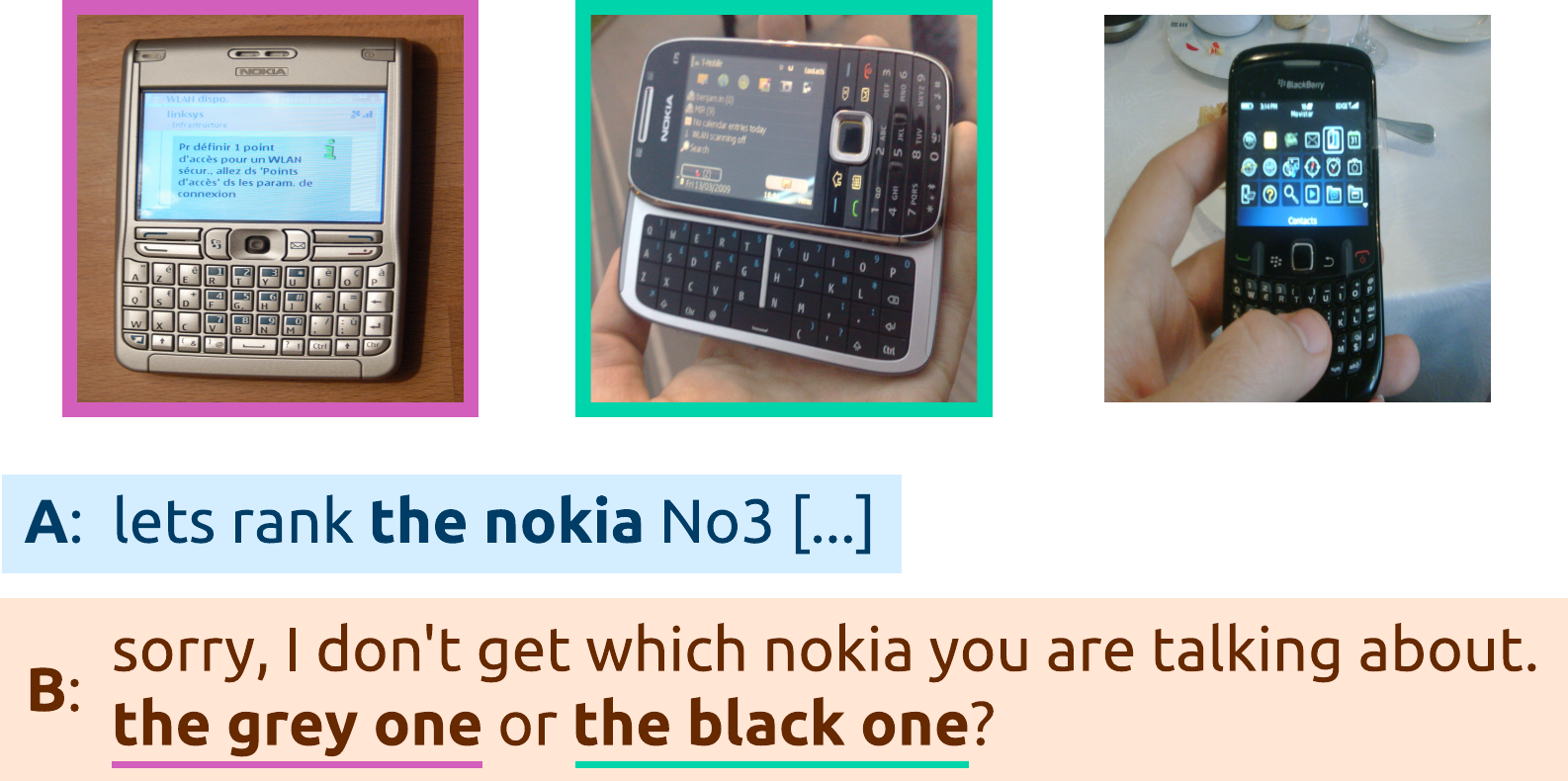}
    \caption{Excerpt (simplified) taken from a dialogue collected by \citet{willemsen_collecting_2022}.
    }
    \label{fig:simplified-dialogue-example}
\end{figure}

A visually grounded dialogue is a conversation in which speakers refer to entities in a (shared) visual context. They do so by producing \textit{referring expressions} (REs). The listener is expected to use the RE to identify the target entity, i.e., the \textit{referent}. Whether the listener is successful in doing so depends on several factors, one being how specific the description of the referent was.
With regard to specification, there exists a trade-off between discriminatory power and efficiency.
On the one hand, the aim is to produce an unambiguous expression with which a referent can be successfully identified, whereas on the other hand a cooperative speaker is expected to make their contribution as economical as possible, while still avoiding ambiguity \citep{grice_logic_1975}.
To illustrate, consider the three phones depicted in Figure \ref{fig:simplified-dialogue-example}.
If the intention of a speaker was to produce a description based on visual content that uniquely identified the phone second from the left, ``\textit{the phone with the QWERTY keyboard}'' would be underspecified, as it applies to both the intended target as well as the right-most image.
To avoid underspecification, additional content could be added to the RE, possibly resulting in a description such as ``\textit{the mostly black Nokia E75 mobile phone with the side-sliding QWERTY keyboard and keypad}''.
This RE does set apart the target from the distractors, but is overspecified, as the description contains more content than is strictly required for identification of the referent in this context, violating the Gricean maxim of quantity \citep{grice_logic_1975}.

In determining form and lexical content of REs, context plays a crucial role.
We will again use Figure~\ref{fig:simplified-dialogue-example} to illustrate this by example.
\textbf{\textsc{A}} attempts to draw the attention of \textbf{\textsc{B}} to a specific phone by referencing its brand name.
However, since \textbf{\textsc{B}} recognizes two phones to be from this brand, \textbf{\textsc{B}} asks a clarification question that focuses on color. 
There are two things to note here. 
First, the REs produced by \textbf{\textsc{B}}, in particular ``\textit{the black one}'', only work as discriminative references due to the mention of the brand name just prior, as ``\textit{one}'' is here a proform of ``\textit{nokia}'' (the right-most phone is also black). 
Second is the symmetry between the REs, showing conventional preservation of form.

For a conversational agent to take part in visually grounded dialogue, it would preferably generate REs in a similar, context-dependent manner, as this is expected by human conversational partners. 
The computational modeling of this process is the domain of referring expression generation (REG), a core natural language generation (NLG) task for which a considerable body of work exists, spanning decades \citep[see e.g.,][]{krahmer_computational_2019}. 
However, REG has traditionally focused primarily on the discriminative properties of REs, leaving discourse-appropriateness in the context of conversation a somewhat understudied problem.

In this paper, we propose an approach to REG for visually grounded dialogue that is meant to satisfy the discriminative property, while simultaneously accounting for discourse-appropriateness. 
We frame the problem as a two-stage process: in the first stage, we model REG as a text- and image-conditioned next-token prediction task: given a dialogue history, i.e., a preceding linguistic context, and the image of a referent, we autoregressively generate an RE as a continuation of the existing linguistic context, using a fine-tuned vision-language model (VLM).
While at this stage we expect to generate an RE that fits the dialogue context and is indicative of the target image, it is not necessarily discriminative with respect to distractors.
We, therefore, propose to use comprehension guiding as part of a \textit{generate-and-rerank} strategy \citep[see e.g.,][]{luo_comprehension-guided_2017} in stage two; our goal being to select an RE with discriminative properties.
Crucially, we introduce \textit{discourse-aware} comprehension guiding as a way to estimate the discriminatory power of candidate REs based on the dialogue context and incorporate this in the candidate selection process.

Our main contributions are as follows:
\begin{itemize}
    \item We propose an approach to REG in visually grounded dialogue based on causal language modeling with multimodal conditioning and fine-tune a generative VLM, here IDEFICS \citep{laurencon_obelics_2023}, for this purpose; 
    \item We show the potential of \textit{discourse-aware} comprehension guiding using the CRDG framework \citep{willemsen_resolving_2023} as part of a modular REG system, with a higher average text-image retrieval accuracy for candidates selected with our reranking schema compared to greedily generated REs according to our human evaluation;
    \item We release the discussed materials, including our LoRA \citep{hu_lora_2022} weights for IDEFICS\footnote{\url{https://github.com/willemsenbram/reg-with-guiding}, \href{https://doi.org/10.5281/zenodo.13225148}{doi:10.5281/zenodo.13225148} \label{fn:zenodo}}.
\end{itemize}

\section{Related work}
REG, as most NLG tasks, has been subject to a paradigm shift over the years.
Whereas earlier methods were mostly symbolic \citep[e.g.,][]{appelt_planning_1985,dale_computational_1995,krahmer_efficient_2002}, most approaches proposed in more recent years are based on neural models \citep[e.g.,][]{mao_generation_2016,luo_comprehension-guided_2017,panagiaris_generating_2021,sun_proposal-free_2023}.
Contemporary NLG research frequently incorporates large language models (LLMs), predominantly those that are Transformer-based \citep{vaswani_attention_2017}.
A common approach to modeling downstream NLG tasks is domain adaptation via transfer learning.
This is typically achieved by fine-tuning a pretrained LLM on a task-specific dataset.

Although the bulk of the computation for most downstream tasks has been delegated to the pretraining of the base model, fine-tuning may still require significant computational resources.
To combat this issue, parameter-efficient fine-tuning methods have been proposed, such as Low-Rank Adaptation \citep[LoRA,][]{hu_lora_2022}.
By freezing the pretrained model weights and instead training rank decomposition matrices that have been added to the dense layers of the network, LoRA manages to reduce the number of trainable parameters by several orders of magnitude, often without considerable adverse effects to downstream performance.

Aside from language, Transformers have shown promising results when it comes to modeling other modalities \citep[e.g.,][]{dosovitskiy_image_2021,radford_robust_2023}. 
Of particular interest here are multimodal models that combine vision and language. 
VLMs such as CLIP \citep{radford_learning_2021} have learned to jointly embed both modalities via contrastive pretraining objectives.
Their learned representations have shown to be useful for discriminative downstream vision-language tasks, such as text-image retrieval (TIR).
We will hereafter refer to these models as discriminative VLMs.
Other VLMs such as Flamingo \citep{alayrac_flamingo_2022}, BLIP-2 \citep{li_blip-2_2023}, Kosmos-2 \citep{peng_grounding_2024}, LLaVA \citep{liu_visual_2023}, and InternVL \citep{chen_internvl_2024} have been introduced to address \textit{generative} downstream tasks, such as image captioning and (multi-turn) visual-question answering. 
These generative VLMs, sometimes called multimodal LLMs (MLLMs), are able to autoregressively output text based on multimodal inputs, as they are built on (pretrained) LLMs with some form of visual input conditioning. 
This makes them particularly useful for inherently multimodal text generation problems such as REG for visually grounded dialogue.

REG has been defined as a task that is chiefly concerned with identification \citep{reiter_building_1997}.
As such, most work in this area emphasizes the discriminative properties of REs.
The goal is to generate an expression with which a referent can be unambiguously identified.
Whether a candidate RE possesses this property is context-dependent, where context represents a multi-faceted concept.

One facet is the visual context in which the referent is embedded, often together with entities that may be mistaken for the referent, i.e., distractors.
Various strategies have been proposed to have neural models take into account the visual context and attempt to maximize discriminatory power of generated REs, including discriminative decoding \citep[e.g.,][]{schuz_decoupling_2021} and comprehension-guiding \citep[e.g.,][]{luo_comprehension-guided_2017}.
These methods typically incorporate some manner of scoring (partial) candidate REs on the basis of their alignment with pragmatic principles, either at inference time to guide decoding, or as part of a \textit{generate-and-rerank} strategy, a commonly used approach for a variety of NLG problems \citep[e.g.,][]{andreas_reasoning_2016,challa_generate_2019,won_break_2023}.
In the latter case, a REG model will generate a set of candidate REs which are reranked on the basis of their discriminatory power according to some referring expression comprehension (REC) model.

These strategies, however, tend to focus primarily on the generation of definite descriptions, disregarding other forms of REs such as pronouns, and do not fully consider the dialogue context in which the REs would be used.
Earlier work on rule-based REG did address some context-sensitive aspects, such as the by \citet{krahmer_efficient_2002} proposed extensions to the influential Incremental Algorithm \citep{dale_computational_1995}, which included reduced descriptions of subsequent mentions and pronominalization. 
More recent work that explicitly considered the linguistic context in addition to the visual context has instead attempted to generate discriminative referring \textit{utterances} \citep{takmaz_refer_2020}, under the assumption, however, that each utterance only mentions a single referent.

\begin{figure*}[ht!]
\centering
\includegraphics[width=1\textwidth]{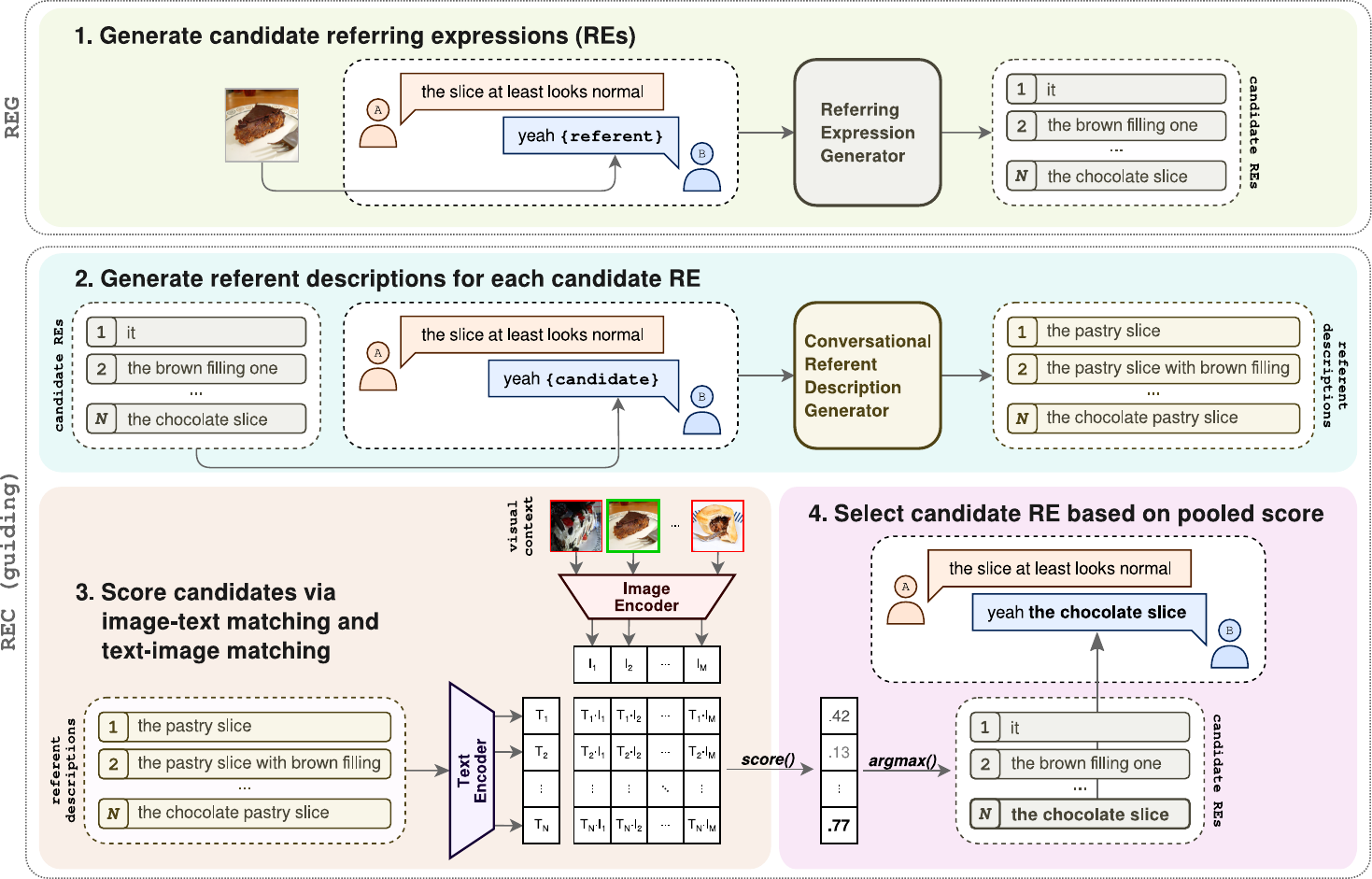}
\caption{
Visualization of the proposed two-stage, four-step framework. The first stage concerns (1) the autoregressive generation of candidate REs where the input to the REG model is the preceding linguistic context of the RE and an image representing the referent. In the second stage, candidate REs are (2) inserted into the dialogue segment at the point at which they were generated, after which the segment is processed by the CRDG \citep{willemsen_resolving_2023} to generate referent descriptions. These referent descriptions are (3) used to evaluate the discourse-dependent discriminatory power of the candidate REs by using a pretrained VLM to produce TIM and ITM scores, which are then (4) weighted to arrive at a composite score for each candidate RE; the highest-scoring candidate RE is selected.}
\label{fig:framework-diagram}
\end{figure*}

\section{Method}
In this work, we focus on generating REs conditioned on a multimodal dialogue context for referents that are represented by independent images. 
This setting bares some resemblance to that of discriminative image captioning \citep[see e.g.,][]{vedantam_context-aware_2017,cohn-gordon_pragmatically_2018,schuz_diversity_2021}.
REG more commonly attempts to describe objects or entities, represented by bounding boxes or segmentation masks, in single images or scenes.
Spatial relations frequently become part of distinguishing descriptions in such settings as a result. 
Our method, however, focuses instead on generating REs based on visual content in situations that have been specifically designed for this to be challenging.
We leave extending the framework to incorporate spatial relations to future work.

\subsection{Task description} 
For a given referent, which is represented by an image (or images), the aim is to generate an RE (1) with which the referent can be identified and (2) which is discourse-appropriate. 

\subsection{Proposed approach}
Broadly speaking, we propose a framework that consists of two components, namely a REG model and a REC model.
For a visualization of this framework, see Figure \ref{fig:framework-diagram}.
We approach REG as a causal language modeling problem.
More specifically, we use a generative VLM that has been pretrained to handle arbitrarily interleaved sequences of text and images \citep{alayrac_flamingo_2022,laurencon_obelics_2023} in order to condition the autoregressive generation of REs on a preceding visio-linguistic context.
For the experiments presented in this paper, the generative VLM we use is IDEFICS \citep{laurencon_obelics_2023}, an open-source implementation of Flamingo \citep{alayrac_flamingo_2022}.
By fine-tuning IDEFICS on visually grounded dialogue data, our aim is to satisfy the second constraint of the task, i.e., generating REs that are a good fit for the projected use context.
In order to ensure the generated REs satisfy the first constraint, we evaluate their discriminatory power using a REC model.
Crucially, as part of a \textit{generate-and-rerank} strategy, we propose \textit{discourse-aware} comprehension guiding.
The motivation for the use of a \textit{discourse-aware} REC model to score discriminatory power comes from the context-dependence of this property, as some REs will need to be resolved to their coreferences in order to be disambiguated and understood to be adequate mentions.
For the experiments presented in this paper, we base our REC model on the conversational referent description generator (CRDG) framework of \citet{willemsen_resolving_2023}.

\subsubsection{Multimodal conditioning with IDEFICS}
IDEFICS is a generative VLM based on the Flamingo VLM architecture \citep{alayrac_flamingo_2022}.
Flamingo was introduced to handle various open-ended vision-language tasks that carry an NLG objective, with a noted focus on using few-shot multimodal in-context learning (ICL) to accomplish them.
Flamingo builds on pretrained vision and language models, bridging these modalities in order to incorporate visual information in the process of predicting the next token.
To condition the autoregressive generation of text on both text and images, gated cross-attention dense layers that are trained from scratch are interleaved between the frozen layers of a pretrained LLM.
Images are encoded using a pretrained vision model, after which the resulting embeddings go through a process of Perceiver-based \citep{jaegle_perceiver_2021} resampling in order to encode the high-dimensional visual feature representations as fixed numbers of so-called visual tokens.
The model cross-attends to this output from the resampler in order to incorporate the visual information into its predictions, enabling the modeling of text interleaved with images.

To use IDEFICS for our purpose, we simply take the available linguistic context, indicating with speaker tokens the identity of the speaker for each message in the dialogue history, and add the image representing the referent to the sequence in the position at which we want to generate an RE.
For reference, see step 1 in Figure \ref{fig:framework-diagram}.

\subsubsection{Comprehension guiding with the CRDG}
\citet{willemsen_resolving_2023} frame reference resolution in visually grounded dialogue as a TIR task.
They note, however, that current discriminative VLMs, typically assume that the text is descriptive of the image. 
As REs in dialogue can take various forms besides definite descriptions, being able to resolve coreferences, including pronouns, is often a prerequisite for successful identification of a referent.
For this reason, they proposed fine-tuning a causal LLM to generate so-called \textit{referent descriptions}. 
Referent descriptions distill all available coreferential information in the linguistic context of a given mention into a single (definite) description of the referent.
These referent descriptions can then be used by a pretrained VLM to identify referents via (zero-shot) TIR.
To illustrate, consider again the REs in Figure \ref{fig:simplified-dialogue-example}. 
If we were to attempt TIR directly with the RE ``\textit{the black one}'', the description is ambiguous, applying to both the target and a distractor.
If we instead use its referent description ``\textit{the black nokia}'', which combines information from all mentions of the referent in the available linguistic context, we now have a distinguishing description.
This shows how the linguistic context is crucially important in resolving an otherwise seemingly underspecified RE and how the CRDG can resolve references regardless of form.

While this framework was originally intended for REC in conversation, we propose to repurpose it as a comprehension-guiding model for REG in visually grounded dialogue.
To evaluate candidate REs generated by our REG model based on their discriminatory power, we insert the candidate RE into the dialogue segment at the position at which it was generated by the REG model, marking its beginning and end in text.
We then use the CRDG to autoregressively generate for this candidate RE a referent description based on the provided dialogue segment.
For reference, see step 2 in Figure \ref{fig:framework-diagram}.
The generated referent description is then encoded with a discriminative VLM to get a text embedding.
We then compute representational similarity between this text embedding and the image embeddings of the candidate referents to rank the candidate REs.
For reference, see step 3 in Figure \ref{fig:framework-diagram}.
Note that the referent descriptions are only used in the process of guiding the selection of candidate REs.

\begingroup
\renewcommand{\arraystretch}{0.9} 
\begin{table}[t!]
\centering
\small
\begin{tabular}{lc}
\toprule
\multicolumn{2}{c}{\textsc{Text-Text}} \\
\midrule
\textbf{Metric} & \textbf{Score}\\
\midrule
BLEU & .71 \\
ROUGE-L & .82 \\
Jaccard & .79 \\
Cosine$_{TT}$ & .92 \\
\bottomrule
\end{tabular}
\begin{tabular}{lc}
\toprule
\multicolumn{2}{c}{\textsc{Text-Image}} \\
\midrule
\textbf{Metric} & \textbf{Score}\\
\midrule
Accuracy & .71 \\
MRR & .83 \\
NDCG & .88 \\
Cosine$_{TI}$ & .48 \\
\bottomrule
\end{tabular}
\caption{\label{table:crdg-ground-truth-results}
Cross-validated performance of incremental version of CRDG framework. Scores are rounded to the nearest hundredth.} 
\end{table}
\endgroup

\noindent\textbf{Candidate reranking}
Although it makes intuitive sense to deem the candidate RE that has the most discriminatory power according to the REC model to be the best available candidate, this is not necessarily always true.
To clarify, consider the following: if we were to simply opt for the candidate RE that has, among the candidates, the highest probability assigned to the target image via softmax, we may be selecting an RE based of a referent description that the VLM considers to be most similar to the target image when accounting for the distractors, but that is not in itself a good description of any of the images. 
Despite low similarity between the images and the description in absolute terms, the relative difference just so happens to be large and in favor of the target image.
As a result, we would likely be selecting a suboptimal RE.

For this reason, we propose to select candidate REs not just based on their \textbf{text$\rightarrow$image} matching (TIM) score, but rerank them based on both their TIM and \textbf{image$\rightarrow$text} matching (ITM) scores: here, the TIM score indicates to what extent the candidate RE describes the target image with respect to the distractor images; the ITM score indicates to what extent the candidate RE describes the target image with respect to the other candidate REs.
Note that each candidate RE is represented by its referent description, as generated by the CRDG, when these scores are computed. 
We combine the scores by way of linear opinion pooling \citep[see e.g.,][]{jacobs_methods_1995}, taking a weighted linear combination of the log softmax of the TIM and ITM logits.
For each candidate RE we calculate its pooled score, $S$, as follows:
\begin{equation*}
\text{$S_i$} = w_{a_i} \cdot \ln(a_i + \varepsilon)  + w_{b_i} \cdot \ln(b_i + \varepsilon) 
\end{equation*}
where, for each $i$-th candidate RE, $a$ and $b$ represent its TIM and ITM softmax probabilities, respectively, each $w$ the coefficient by which $a$ and $b$ are scaled, and $\varepsilon$ a small constant that is added to avoid taking the (theoretical) log of $0$.  
The coefficients sum to $1$.
We select the candidate RE with the highest $S$ for the target image\footnote{Although we only consider the output from a single VLM here, it is possible to aggregate scores from multiple VLMs, treating each as an independent ``expert''. Moreover, in addition to the VLM-based TIM and ITM scores, other properties of interest may also be incorporated as (weighted) ``opinions''.}.
We describe a hypothetical case in Appendix \ref{sec:appendix-reranking} to further illustrate the rationale behind this weighted reranking.

\begingroup
\renewcommand{\arraystretch}{0.9} 
\begin{table*}[t!]
\centering
\footnotesize
\begin{tabular}{lccccccc}\toprule
& \multicolumn{3}{c}{\textsc{Text-Text}} & \multicolumn{4}{c}{\textsc{Text-Image}}
\\\cmidrule(lr){2-4}\cmidrule(lr){5-8}
& \textbf{BLEU} & \textbf{ROUGE-L} & \textbf{Cosine$_{TT}$} & \textbf{Accuracy} & \textbf{MRR} & \textbf{NDCG} & \textbf{Cosine$_{TI}$} \\\midrule
$1$-shot & .30 & .34 & .64 & .57 & .74 & .80 & .47 \\
$2$-shot & .32 & .36 & .65 & .58 & .74 & .81 & .47 \\
$4$-shot & .32 & .35 & .64 & .53 & .71 & .78 & .46 \\
$8$-shot & .31 & .34 & .64 & .49 & .67 & .76 & .45 \\\midrule
FT & .40 & .48 & .72 & .67 & .81 & .86 & .48 \\\bottomrule
\end{tabular}
\caption{\label{table:idefics-greedy-results}
Cross-validated \textit{n}-shot and fine-tuned (FT) REG performance of IDEFICS using greedy decoding. Text generation metrics use \textit{ground truth} REs as reference. Scores for TIR metrics are based on generated referent descriptions. Scores are rounded to the nearest hundredth.}
\end{table*}
\endgroup

\section{Experiments}

\subsection{Data}
The dialogues used in our experiments come from the visually grounded dialogue task A Game Of Sorts \citep[AGOS,][]{willemsen_collecting_2022}.
In this ``game'', two players are presented with a set of nine images that they are asked to rank---one at a time---based on a given sorting criterion. 
To complete the task, they will have to agree on a ranking which they deem satisfactory.
The game is played over multiple rounds with the same set of images to ensure repeated mentions of the same referents. 
Although the players see the same set of images, they cannot see each other's perspective. 
The position of the nine images on screen is randomized, forcing the players to refer to the images based on their visual content.
The task was specifically designed to encourage discussions and imposes no restrictions on message content.
As a result, the referring language comes embedded in considerably longer and more diverse conversations compared to those from related work.
\citet{willemsen_collecting_2022} collected 15 dialogues in total: three dialogues for each one of five image categories.
Images from the same set were selected to have overlapping visual attributes, in order to further complicate the production of discriminative REs. 
Due to the deliberate challenges to the referential process and the relatively unconstrained nature of the dialogues, the task can be considered a challenging test bed for the grounding and generation of REs in conversation.

For fine-tuning and evaluation of both REG and REC models, we require dialogues with REs annotated.
For this purpose, we use the span-based mention annotations for AGOS from \citet{willemsen_resolving_2023}. 
These annotations indicate the start and end of all the mention spans found in the dialogues, and the image, or images, to which they refer.
We will consider these human-produced REs to be the \textit{ground truth} for our study. 

\subsection{Evaluation}
We focus on evaluating single-image referents, however noting that, in principle, our proposed framework can be extended to the multi-image referent case.
We adopt the cross-validation protocol used by \citet{willemsen_resolving_2023}, where the AGOS dataset is partitioned along the five image sets: for each run, twelve dialogues from four image sets are used for training, and the three dialogues of the remaining image set are used for testing.
We limit the context window of the dialogue to the previous seven messages for model-based experiments, and report TIR results based on the reduced visual context, i.e., not considering ranked images to be part of the candidate referents.

\begingroup
\renewcommand{\arraystretch}{0.9} 
\begin{table*}[t!]
\centering
\footnotesize
\begin{tabular}{lccccccc}\toprule
& \multicolumn{3}{c}{\textsc{Text-Text}} & \multicolumn{4}{c}{\textsc{Text-Image}}
\\\cmidrule(lr){2-4}\cmidrule(lr){5-8}
& \textbf{BLEU} & \textbf{ROUGE-L} & \textbf{Cosine$_{TT}$} & \textbf{Accuracy} & \textbf{MRR} & \textbf{NDCG} & \textbf{Cosine$_{TI}$} \\\midrule
Top-1 & .21 & .41 & .71 & .60 & .76 & .82 & .47 \\
Max disc. & .29 & .40 & .70 & .89 & .94 & .95 & .50 \\
Rerank & .31 & .40 & .70 & .86 & .92 & .94 & .51 \\\bottomrule
\end{tabular}
\caption{\label{table:idefics-beam-results}
Cross-validated REG performance of fine-tuned IDEFICS using beam search decoding with a width of $6$. Text generation metrics use \textit{ground truth} REs as reference. Scores for TIR metrics are based on generated referent descriptions. Scores are rounded to the nearest hundredth.}
\end{table*}
\endgroup

\subsubsection{Metrics}
We score the referent descriptions generated by the CRDG based on their similarity to the manually constructed ground truth labels using the same metrics as reported in \citet{willemsen_resolving_2023}, i.e., the Jaccard index, BLEU (based on unigrams and bigrams) \citep{papineni_bleu_2002}, ROUGE-L \citep{lin_rouge_2004}, and cosine similarity between text embeddings (Cosine$_{TT}$). 
When comparing generated REs against ground truth mentions, we compute unigram-based BLEU, ROUGE-L, and cosine similarity between text embeddings (Cosine$_{TT}$)\footnote{Note that metrics based on overlapping content are not as robust for more open-ended tasks such as REG; we consider them here as secondary indicators for model selection.}.
We report TIR performance in terms of top-1 accuracy, mean reciprocal rank (MRR), normalized discounted cumulative gain (NDCG), and cosine similarity between referent description text embeddings and target image embeddings (Cosine$_{TI}$).
Model-based TIR results reflect the zero-shot performance of the discriminative VLM as it is used in the CRDG framework.
This VLM is also used to get the embeddings for the cosine similarity measures.
All metrics are bound between $[0,1]$.

\subsubsection{Human}
In order to externally validate our model-based experimental results, we conduct a human subjects experiment to evaluate human TIR performance for generated REs and to compare these results to those for the ground truth.
Following \citet{willemsen_resolving_2023}, participants are shown the REs in the context of the unfolding dialogue. We, however, show the dialogue up until the end of the current RE for which the participant is asked to provide an answer.
We evaluate with the reduced visual context.
For more details, see Appendix \ref{sec:appendix-human-eval}.

\subsection{Comparisons}
Given the focus on multimodal ICL with Flamingo \citep{alayrac_flamingo_2022}, we evaluate the \textit{n}-shot performance of IDEFICS in addition to its (LoRA) fine-tuned performance.
We compare these variants based on outputs generated using greedy decoding.
For details about the selection of support examples for ICL, see Appendix \ref{sec:support-examples}.
Further experiments use the fine-tuned variants of the model. 
To generate multiple candidate REs, we use beam search with a width of $6$.
We examine how our proposed approach using weighted reranking (Rerank), which selects candidates based on their pooled score, compares against ablated versions of the method. We contrast performance with a variant that selects the candidate with the most discriminatory power (Max disc.) and a variant without any guiding that simply selects the top beam hypothesis (Top-1). 
We deliberately focus on evaluating different versions of the proposed framework, as, to the best of our knowledge, existing REG models are ill-suited to handle the AGOS task setting or principally do not satisfy our discourse-appropriateness criterion.
For instance, if we were to use as a baseline a model that would invariably generate context-independent, but overspecified or caption-like REs---such as discussed in Section \ref{sec:introduction} in relation to the example based around Figure \ref{fig:simplified-dialogue-example}---these may result in high TIR accuracy, but, even so, will virtually never be discourse-appropriate.

\subsection{Implementation details}
Similar to \citet{willemsen_resolving_2023}, we obtain the CRDG by fine-tuning GPT-3---although davinci-002 instead of the davinci base model---using the OpenAI API.
Crucially, however, our version of the CRDG is incremental as opposed to message-based.
We use InternVL \citep{chen_internvl_2024}, specifically InternVL-G, as our discriminative VLM within the CRDG framework.
With regard to the reranking of candidate REs, although we could treat the coefficients as learnable parameters, we instead simply set $w$ to $\frac{2}{3}$ and $\frac{1}{3}$ for the TIM and ITM scores, respectively, as we believed this to represent a reasonable trade-off between the scores for our purpose.
All experiments reported in this paper that involve IDEFICS are based on the 80 billion parameter variant\footnote{\url{https://huggingface.co/HuggingFaceM4/idefics-80b}}.
We use quantized LoRA \citep[QLoRA,][]{dettmers_qlora_2023} for parameter-efficient fine-tuning.
We modify the loss calculation by masking the loss for all tokens but the RE.
We estimate, without exhaustive search, hyperparameters for IDEFICS fine-tuning using nested five-fold cross-validation. 
For additional details, including IDEFICS and GPT-3 hyperparameters, see Appendix \ref{sec:appendix-implementation-details}.

\section{Results}
Our results are based on 1305 of the 1319 annotated mentions of single-image referents; 14 samples were excluded as their target referents were not part of the set of candidate referents as a consequence of evaluating with the reduced visual context.
Table \ref{table:re-examples} shows REs from different sources for a few dialogue samples.

\begingroup
\renewcommand{\arraystretch}{0.9} 
\begin{table}[t!]
\centering
\footnotesize
\begin{tabular}{lc}\toprule
& \textbf{Accuracy} \\\midrule
Greedy & .74  \\
Rerank & .78  \\
Ground truth & .88  \\\bottomrule
\end{tabular}
\caption{\label{table:human-eval-experiment-results}
Human (incremental) reference resolution performance. Scores are rounded to the nearest hundredth.}
\end{table}
\endgroup

\noindent\textbf{Incremental CRDG} Table \ref{table:crdg-ground-truth-results} shows the performance of the CRDG on the ground truth data. 
We managed to closely replicate the results reported by \citet{willemsen_resolving_2023} despite our variant of the CRDG being incremental.

\noindent\textbf{Multimodal ICL vs. fine-tuning} In Table \ref{table:idefics-greedy-results} we show results for candidate REs generated using greedy decoding with $1$-, $2$-, $4$-, and $8$-shot multimodal ICL and with the fine-tuned model.
We found that a single example tended to be enough for the model to generate an RE, in accordance with the provided task.
Adding an additional example improved performance slightly, but further increasing the number of support examples hurt performance instead.
Moreover, the metrics showed a notable gap between ICL and fine-tuning, with fine-tuning averaging higher scores across the board.

\noindent\textbf{Ablations} Shown in Table \ref{table:idefics-beam-results} are results of the three strategies for candidate selection after beam search.
With the exception of text-image cosine similarity, we observed slightly lower scores for the TIR metrics for the reranked REs in comparison with those that had the most discriminatory power. 
This was expected, as we actively went against taking the most discriminative candidate with our weighted reranking, which, our results suggested, did lead to higher representational similarity, on average, between referent descriptions and target images. 
These differences were, however, marginal.

\noindent\textbf{Human performance} We validated our model-based experimental results through human evaluation, results of which are shown in Table \ref{table:human-eval-experiment-results}.
We collected one data point per dialogue, meaning 15 data points per source of RE listed, for a total of 45 data points from 38 different participants.
We contrasted TIR accuracy for REs generated with fine-tuned IDEFICS with that of ground truth mentions.
We found that, although lagging behind the ground truth, the generated REs, regardless of the exact strategy, showed strong performance, far exceeding chance level (which was roughly 22\%).
Although both tested model-based RE variants seemed effective, our reranked REs resulted in higher accuracy than those based on greedy decoding.

\noindent\textbf{RE length} We found that REs generated by our (fine-tuned) REG model tend to be shorter, on average, than the ground truth mentions. 
This is one indicator of our model not having been prone to generating overspecified REs, which would otherwise have had the potential to artificially inflate accuracy scores.
A comparison between the average length of the generated REs and the ground truth is visualized in Figure \ref{fig:average-re-length} in Appendix \ref{sec:appendix-results}.

\noindent\textbf{RE content} When examining the ground truth REs, we found that more than 20 percent of the included mentions contain no words that were descriptive of visual content (e.g., \textit{``it''}, \textit{``that one''}), with the pronoun \textit{``it''} accounting for roughly half of these REs. 
We found that such REs were selected at a similar rate when using our weighted reranking schema.
It is worth nothing, however, that whenever both the ground truth and selected candidate REs contained no content words, their forms would, at times, differ (e.g., \textit{``it''} having been selected where the ground truth was \textit{``that one''}).

\begingroup
\renewcommand{\arraystretch}{1} 
\begin{table*}[t!]
\centering
\footnotesize
\begin{tabular}{l|l|l|l}\toprule
\rotatebox[origin=c]{90}{\textsc{Visual context}} & \makecell{\includegraphics[width=0.25\linewidth]{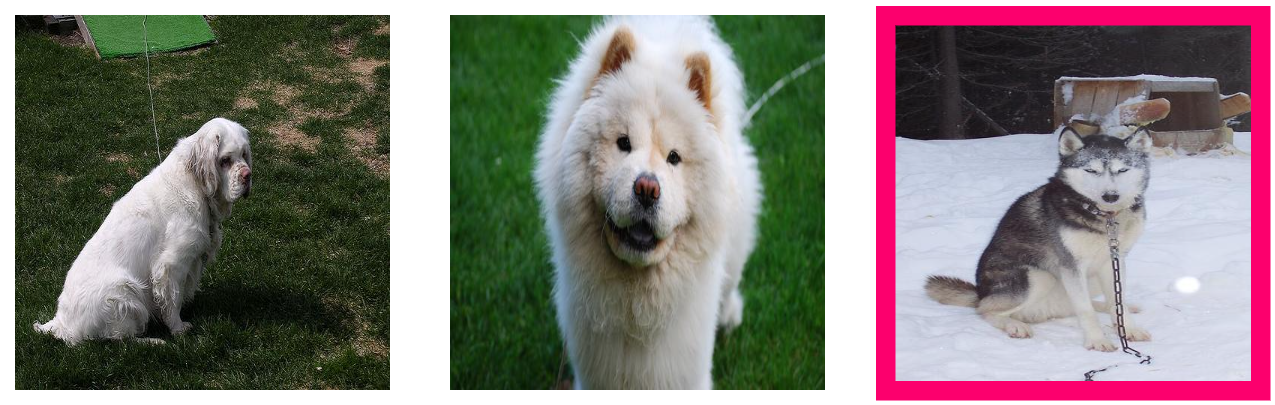}} & \makecell{\includegraphics[width=0.25\linewidth]{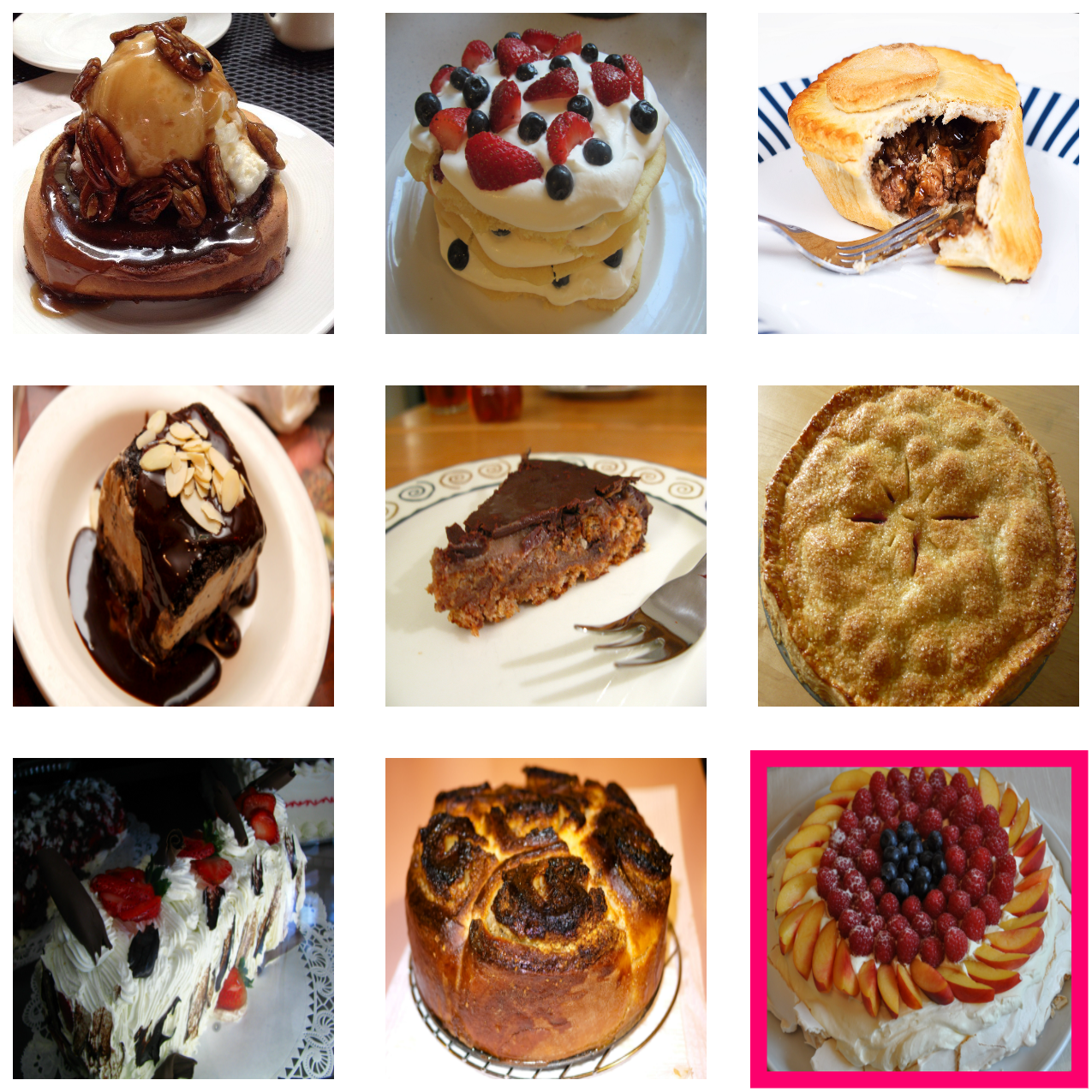}} & \makecell{\includegraphics[width=0.17\linewidth]{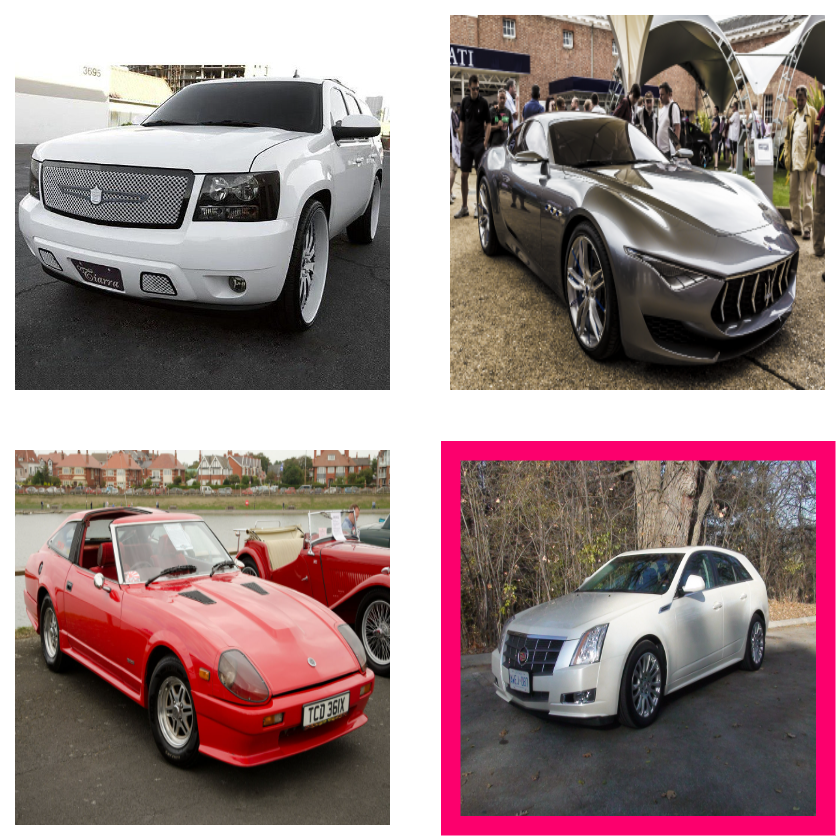}} \\
\midrule 
\multirow{10}{*}{\rotatebox[origin=c]{90}{\textsc{Linguistic context}}} & \multirow{2}{*}{\textcolor{message_gray}{\makecell{[...]}}} & \multirow{2}{*}{\textcolor{message_gray}{\makecell{[...]}}} & \multirow{2}{*}{\textcolor{message_gray}{\makecell{[...]}}} \\ 
& & & \\
& \multirow{3}{*}{\textcolor{message_blue}{\makecell[l]{\textbf{A}: The poodle is the one that \\ looks like a sheep right?}}} & \multirow{3}{*}{\textcolor{message_blue}{\makecell[l]{\textbf{A}: the chocolate one now maybe? \\ at least it has no cream, and some \\ nuts}}} & \multirow{3}{*}{\textcolor{message_blue}{\makecell[l]{\textbf{A}: didnt we say the white suv \\ was more solid than grey and red?}}} \\ 
 & & & \\
 & & & \\
 & \multirow{1}{*}{\textcolor{message_orange}{\makecell{\textbf{B}: yeah}}} & \multirow{1}{*}{\textcolor{message_orange}{\makecell[l]{\textbf{B}: ah true I didn't see the nuts there}}} & \multirow{1}{*}{\textcolor{message_orange}{\makecell{\textbf{B}: red then}}} \\ 
& \multirow{2}{*}{\textcolor{message_orange}{\makecell{\textbf{B}: and now the husky}}} & \multirow{2}{*}{\textcolor{message_blue}{\makecell[l]{\textbf{A}: I'm not sure if it is ice cream to \\be honest}}} & \multirow{2}{*}{\textcolor{message_blue}{\makecell{\textbf{A}: but sure we can swap}}} \\ 
& & & \\ 
& \multirow{2}{*}{\makecell{\textcolor{message_blue}{\textbf{A}: Husky is} \texttt{\textbf{\{RE\}}} \textcolor{message_gray}{right?}}} & \multirow{2}{*}{\makecell[l]{\textcolor{message_orange}{\textbf{B}: The round one with lots of fruit?} \\ \texttt{\textbf{\{RE\}}}\textcolor{message_gray}{'s big and beautiful}}} & \multirow{2}{*}{\makecell{\textcolor{message_blue}{\textbf{A}:} \texttt{\textbf{\{RE\}}} \textcolor{message_gray}{now?}}} \\ 
& & & \\ \midrule 
\textbf{Greedy} & the one with the chain & It & white \\
\textbf{Top-1} & it & It & white \\
\textbf{Max disc.} & it & It & white sedan \\
\textbf{Rerank} & the one with the chain & It & white sedan \\
\textbf{GT} & the one with a chain in the snow & It & white suv \\
\bottomrule
\end{tabular}
\caption{\label{table:re-examples}
Examples of REs as produced by different versions of the proposed method, all generated with fine-tuned IDEFICS. \textbf{Greedy} shows REs generated using greedy decoding, \textbf{Top-1} means REs that were the top beam search result, \textbf{Max disc.} are REs generated with beam search that had the most discriminatory power, and \textbf{Rerank} are REs that were selected based on our weighted reranking. Also shown are the \textit{ground truth} (\textbf{GT}) REs. The \textsc{Visual context} depicts, for each dialogue, the unranked images at the time the ground truth RE was produced; the target referent is highlighted (magenta-colored border around the image). The \textsc{Linguistic context} shows (a limited number of) the preceding messages and the current message up until the start of the RE (\texttt{\textbf{\{RE\}}}); the light-gray text shows the remainder of the original message after the RE.}
\end{table*}
\endgroup

\section{Discussion}
In this paper, we explored the problem of REG in visually grounded dialogue.
Our aim was to realize the generation of REs that were not only discriminative, but also appropriate for the dialogue context in which they would be used.
We proposed to approach the problem from a causal language modeling perspective, where the generation of tokens would be conditioned on both text and images.
By fine-tuning a generative VLM, IDEFICS \citep{laurencon_obelics_2023}, we showed it is possible to generate REs that are indicative of the referent while suitable for the dialogue context.
Notably, we were successful using a parameter-efficient fine-tuning approach \citep{dettmers_qlora_2023} and while having relatively limited data for training \citep{willemsen_collecting_2022}.
In addition, we introduced \textit{discourse-aware} comprehension-guiding to evaluate whether candidate REs are discriminative given their linguistic context.
By adding candidate REs to the dialogue for which they were generated, we were able to use the CRDG framework of \citet{willemsen_resolving_2023} to score candidate REs on their discourse-dependent discriminatory power.
Finally, we showed that human TIR accuracy using candidate REs selected based on a weighted reranking of scores derived from this discourse-aware REC model was on average higher than for candidate REs generated through greedy decoding.

One of the main benefits of our approach is the ability for the REG model to generate REs that are commonly used in dialogue, but for which discriminatory power is neigh impossible to estimate without having an understanding of preceding linguistic context.
A typical example of such REs are pronouns.
As a result of our REC model being discourse-aware, our REG model is free to generate pronouns and other constructions involving proforms if these are deemed probable continuations of the current linguistic context, as the REC model will be able to evaluate whether these candidate REs are, in fact, discriminative.

With respect to the human evaluation, what is notable is that the model-based REs were generated based on a limited context window that included only the seven previous messages. 
The ground truth mentions, logically, were produced while the speakers had access to and knowledge of the entire dialogue history, the linguistic as well as the extralinguistic context.
By evaluating using the unfolding dialogues in their entirety instead of limiting these to a rolling window of eight messages, we biased the human evaluation slightly towards the ground truth; this was a conscious design choice as not doing so would unfavorably bias results towards the models instead.
In light of this, our results are arguably even more promising.

Furthermore, rather than incorporating the entire visual context, our REG model was only conditioned on an image of the referent when generating an RE.
As a result, the generated REs were generally descriptive, but not necessarily discriminative.
Although we have now relied on our REC model to filter out such candidates, we suggest future research to consider the possibility of improving the generated candidates in terms of their discriminatory power by including the visual context as part of the input to the REG model.
Related, we suggest testing alternative decoding strategies, for example those that are sampling-based or, perhaps more appropriate, ones that aim to be discriminative \citep[e.g.,][]{schuz_decoupling_2021}.

\section*{Limitations}
The experiments reported in this paper were based solely around modeling the English language; it is of yet unclear whether our results would transfer to other languages.
We have focused on a single, relatively small dataset for which the annotations required by our approach were available; acquiring similar annotations for other, bigger datasets would be relatively costly.
We have experimented with only one generative VLM for this paper; as a result, we do not know to what extent our findings generalize to other generative VLMs.
We have used a closed-source API-based method for fine-tuning of the CRDG; consequently, we are not able to make the model weights publicly available, nor is the fine-tuning process transparent.
The current iteration of the CRDG is unimodal, whereas the task of resolving references in visually grounded dialogue is inherently multimodal; this limits the maximally achievable performance.
Our approach is modular and, as such, likely to be affected by error propagation; a bottleneck is the CRDG framework if it overvalues inadequate candidates (false positives) or undervalues adequate ones (false negatives) with respect to their discriminatory power.
We currently operate on the assumption that utterance planning has been delegated to another system; this is a complex problem and challenging to solve properly, but will likely ultimately require a more unified approach that implicitly includes REG.

\section*{Acknowledgements}
This work was partially supported by the Wallenberg AI, Autonomous Systems and Software Program (WASP) funded by the Knut and Alice Wallenberg Foundation. The authors would like to thank Dmytro Kalpakchi, Jim O'Regan, Travis Wiltshire, Chris Emmery, and the anonymous reviewers for their helpful comments.

\bibliography{references}

\appendix

\section{Reranking}
\label{sec:appendix-reranking}

\begin{figure}[ht!]
\centering
\includegraphics[width=0.95\linewidth]{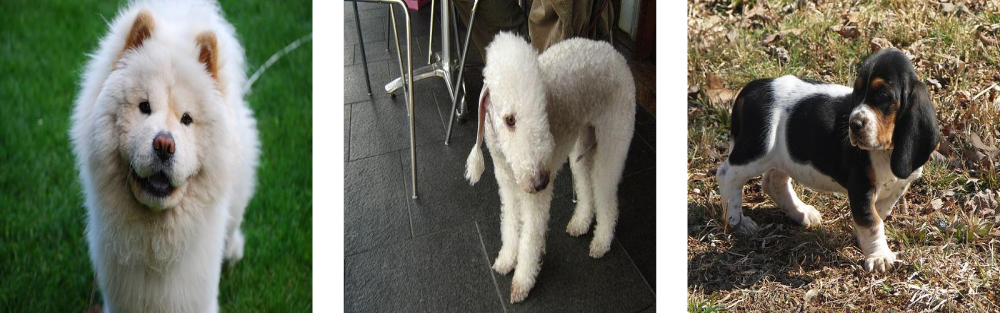}
\caption{Images of dogs for the example in Appendix \ref{sec:appendix-reranking} to illustrate the rationale behind weighted reranking.}
\label{fig:dogs-ranking-example}
\end{figure}

We will further illustrate the need for reranking using a simplified, hypothetical example based around the images in Figure \ref{fig:dogs-ranking-example}.
Figure \ref{fig:dogs-ranking-example} depicts three images of dogs. 
We will consider the left-most image to be our target, with the other two serving as distractors.
We have three candidate REs for the target image: \textit{``the white dog''}, \textit{``the green car''}, and \textit{``the attentive dog''}.
Of these three candidates, \textit{``the attentive dog''} is arguably the most appropriate. 
The RE \textit{``the green car''} does not fit the target image nor does it describe the distractors, as none depict a car. 
The RE \textit{``the white dog''} is underspecified, as it applies to both the target image and a distractor (the middle image). 
Given that the target image depicts a dog that looked directly at the camera when its picture was taken, which is not true for the other dogs, using the adjective \textit{``attentive''} should be acceptable.

Now, in order to perform candidate selection, we use a discriminative VLM to encode each candidate RE and each image that is part of the visual context.
If we then compute representational similarity between text and image embeddings, followed by a softmax over the resulting logits per candidate RE, we get what we consider a probability distribution over the images per candidate RE. 
This is expected to provide some indication with respect to how well the target image is described by each candidate RE given the current visual context.

However, in the scenario that we have sketched here, the following may happen.
Although \textit{``the green car''} has low representational similarity in absolute terms with each image, due to the greater presence of the color green in the target image it scores considerably higher than the distractor images for this candidate RE, which is amplified by the application of the softmax function.
As a result, in this hypothetical, the softmax score for the target image for the candidate RE \textit{``the green car''} would be considerably higher than the score of the more appropriate \textit{``the attentive dog''}. 
Clearly, selecting REs based solely on this score is not appropriate.

One way to address this is to not only apply the softmax over the images per candidate RE, but to also apply it over the candidate REs for the target image.
This will provide an indication for how well the target image is described by each candidate RE, in relation to the other candidates.
The highest softmax score is likely assigned to \textit{``the white dog''}, with \textit{``the attentive dog''} in close second, and \textit{``the green car''} a distant third.
The candidate \textit{``the white dog''} would be an acceptable RE were it not for the fact that it also applies to a distractor.
If we were to select REs based solely on this score, we are more likely to select a candidate that is descriptive, but not discriminative.

Thus, we instead combine the two scores to arrive at a composite that more accurately represents the appropriateness of the candidate REs in the given context than each score independently would.
We gain further control over the trade-off between descriptive and discriminative through weighting.

\section{Human evaluation}
\label{sec:appendix-human-eval}
\begin{figure*}[h!]
\centering
\includegraphics[width=1\textwidth]{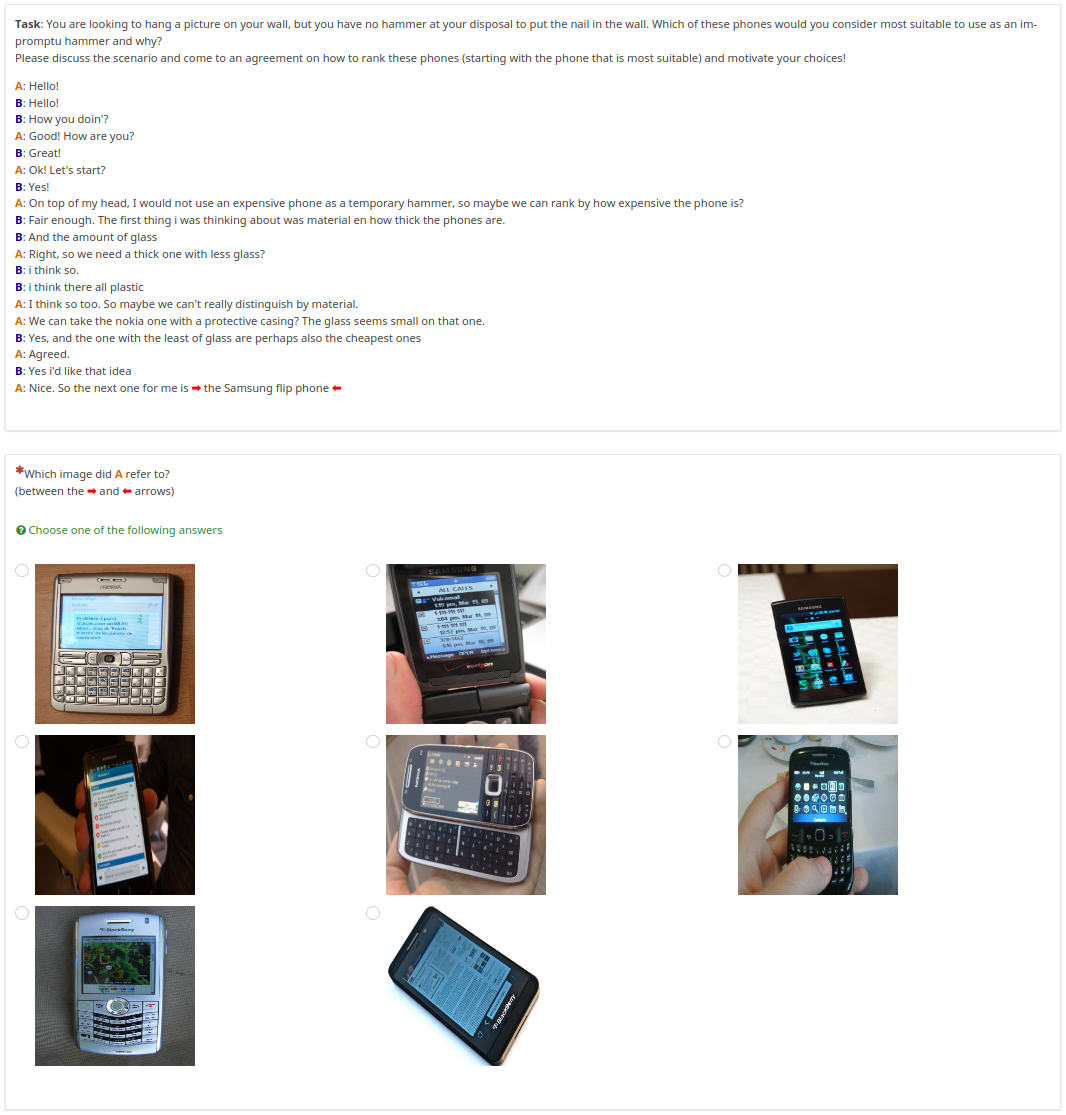}
\caption{Example of an item shown to participants during the human evaluation study.}
\label{fig:human-eval-experiment-item}
\end{figure*}

\begin{figure*}[h!]
\centering
\includegraphics[width=1\textwidth]{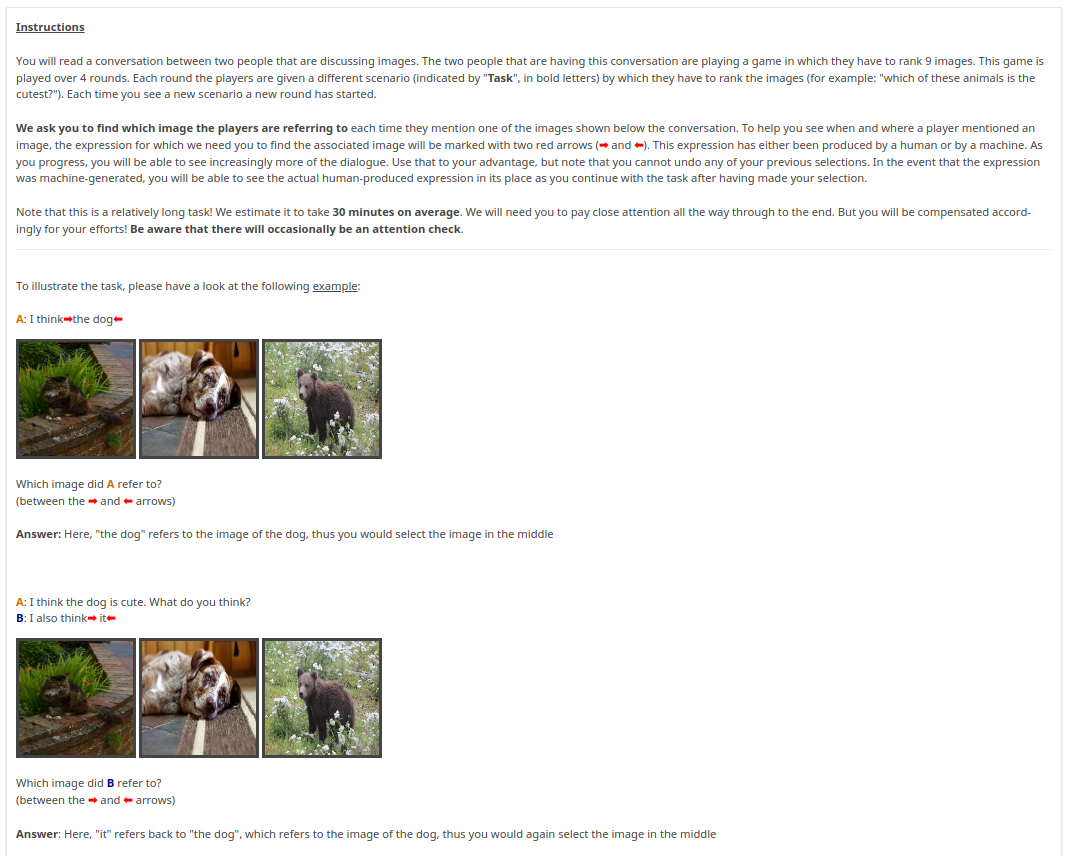}
\caption{Instructions as shown to participants during the human evaluation study (1/2).}
\label{fig:human-eval-experiment-instructions-1}
\end{figure*}

\begin{figure*}[h!]
\centering
\includegraphics[width=1\textwidth]{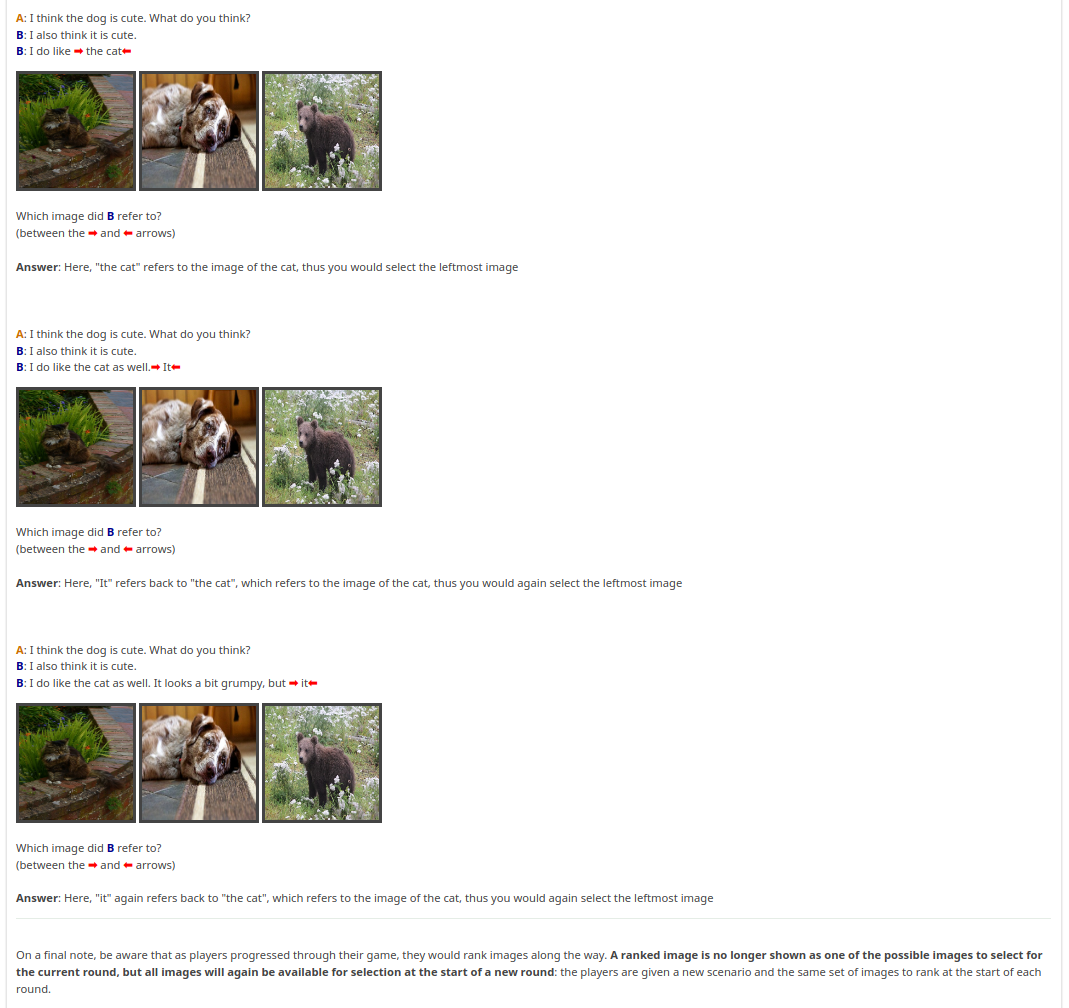}
\caption{Instructions as shown to participants during the human evaluation study (2/2).}
\label{fig:human-eval-experiment-instructions-2}
\end{figure*}

\begin{figure*}[h!]
\centering
\includegraphics[width=1\textwidth]{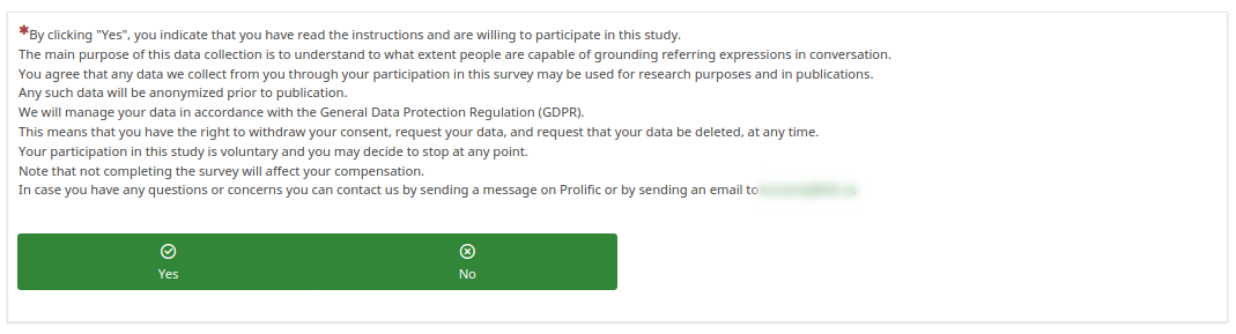}
\caption{Participant informed consent for human evaluation study.}
\label{fig:human-eval-experiment-consent}
\end{figure*}

Instructions provided to participants are shown in Figure \ref{fig:human-eval-experiment-instructions-1} and Figure \ref{fig:human-eval-experiment-instructions-2}, with the informed consent question shown in Figure \ref{fig:human-eval-experiment-consent}.
An example of a task-related question is shown in Figure \ref{fig:human-eval-experiment-item}.
The order of the images is randomized per question.
An attention check is added after every 25 task-related questions.
The survey platform we used was LimeSurvey\footnote{\url{https://www.limesurvey.org/}}, with participants recruited via Prolific\footnote{\url{https://www.prolific.com/}}.
Eligible workers had a minimum approval rate of 99\%, a minimum of 500 previously completed submissions, and had indicated that they are fluent in English.
Regardless of the source of the RE, the participants were allowed to provide data for at most one dialogue per image set.
The expected time-on-task was adjusted based on the number of questions, which varied due to a variable number of REs per dialogue.
Participants were financially compensated for their contributions, with compensation affected by the expected time-on-task.

\section{Support examples}
\label{sec:support-examples}
In order to select suitable support examples for multimodal ICL, we examined the dialogues to find the most frequently occurring forms of REs. 
We identified four categories of REs for which we selected two support examples per image category.
The RE categories were (in)definite descriptions (e.g., \textit{``the white curly dog''}), pronouns (e.g., \textit{``it''}), noun phrases that included a proform in addition to content words (e.g., \textit{``the black one''}), and noun phrases that contained no content words (e.g., \textit{``that one''}).
They are listed here in order of importance, meaning for 1-shot ICL the support example was taken from the (in)definite descriptions category, 2-shot had a support example for both the (in)definite descriptions and pronouns categories, and so on.
For each support example we added the preceding seven messages from the dialogue history and the (partial) task description that was shown to the participants.
Examples were formatted according to the ``User-Assistant'' template, where the ``User'' provides the dialogue segment up until the start of the RE and the ``Assistant'' provides the RE in response.

\section{Additional implementation details}
\label{sec:appendix-implementation-details}
For both fine-tuning and inference, we distribute the model over 8 x 24GB NVIDIA GeForce RTX 3090 using naive model parallelism.
Hyperparameters for IDEFICS fine-tuning are provided in Table \ref{table:hyperparams-idefics-fine-tuning}.
Hyperparameters for GPT-3 fine-tuning via the OpenAI API are provided in Table \ref{table:hyperparams-gpt3-fine-tuning}.

Training samples for IDEFICS fine-tuning were formatted as follows: 

\begingroup
\begin{verbatim}
  [bos token] +
  [preceding linguistic context] + 
  [referent image] + 
  [start of RE token] + 
  [RE] +
  [end of RE token] +
  [eos token]
\end{verbatim}
\endgroup

\noindent Note that the preceding linguistic context included a (partial) task description.
Separate messages were joined by newline characters.
The following is an example of a sample (shortened window for illustrative purposes): \\ {\color{purple}\verb|<s>| M: Your neighbour's cat frequently uses your garden as its own personal bathroom. You decide to adopt a dog to deal with this issue. Which of these dogs would be most effective in scaring off the neighbour's cat and why?}{\color{blue}\verb|\n|}{\color{purple}A: yeah lets go for chow}{\color{blue}\verb|\n|}{\color{purple}B: And then \verb|<referent_image>| \verb|>>| the husky \verb|<<| \verb|</s>|}

\begingroup
\renewcommand{\arraystretch}{0.9} 
\begin{table}[th!]
\centering
\footnotesize
\begin{tabular}{lc}\toprule
Epochs                      & 1    \\
Batch size                  & 1    \\
Gradient accumulation steps & 4    \\
Learning rate               & 7e-5 \\
\midrule
LoRA $r$                      & 16   \\
LoRA $\alpha$                  & 32   \\
LoRA dropout                & 0.1  \\   
\bottomrule
\end{tabular}
\caption{Hyperparameters for fine-tuning of IDEFICS-80b. We use default values if not otherwise specified.}
\label{table:hyperparams-idefics-fine-tuning}
\end{table}
\endgroup

\begingroup
\renewcommand{\arraystretch}{0.9} 
\begin{table}[th!]
\centering
\footnotesize
\begin{tabular}{lc}\toprule
Epochs                      & 3 \\
Batch size                  & 2 \\
Learning rate multiplier    & 2 \\
\bottomrule
\end{tabular}
\caption{Available hyperparameters for fine-tuning of GPT-3 (davinci-002) using the OpenAI API.}
\label{table:hyperparams-gpt3-fine-tuning}
\end{table}
\endgroup

\section{Additional results}
\label{sec:appendix-results}

\begin{figure}[ht!]
\centering
\includegraphics[width=0.99\linewidth]{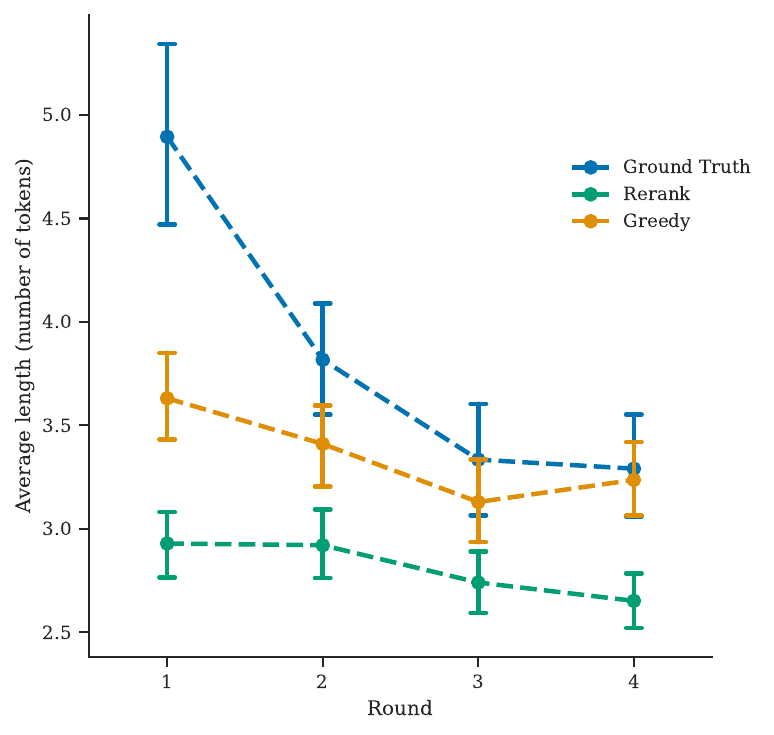}
\caption{Average RE length per round. Shown are \textit{ground truth} REs taken from the dialogues (blue), REs generated by the fine-tuned IDEFICS model using greedy decoding (orange), and REs selected based on our weighted reranking (green). Error bars indicate 95\% bootstrapped confidence intervals.}
\label{fig:average-re-length}
\end{figure}

\end{document}